%% file: main_version6_arxiv_0411.tex
\documentclass[letterpaper,journal]{IEEEtran}
\usepackage{xcolor} 
\usepackage{amsmath,amsfonts}
\usepackage{array}
\usepackage[caption=false,font=normalsize,labelfont=sf,textfont=sf]{subfig}
\usepackage{textcomp}
\usepackage{stfloats}
\usepackage{url}
\usepackage{verbatim}
\usepackage{graphicx}
\usepackage{algpseudocode}

\usepackage[numbers,sort&compress]{natbib}
\usepackage[colorlinks=true,linkcolor=blue,citecolor=blue,urlcolor=blue]{hyperref}

\makeatletter
\renewcommand\NAT@open{\textcolor{blue}{[}}
\renewcommand\NAT@close{\textcolor{blue}{]}}
\makeatother

\def\BibTeX{{\rm B\kern-.05em{\sc i\kern-.025em b}\kern-.08em
    T\kern-.1667em\lower.7ex\hbox{E}\kern-.125emX}}
\usepackage{balance}

\begin{document}

\title{\LARGE \bf
Look-to-Touch: A Vision-Enhanced Proximity and Tactile Sensor for Distance and Geometry Perception in Robotic Manipulation
}

\author{
    \IEEEauthorblockN{
        Yueshi Dong,
        Jieji Ren,
        Zhenle Liu,
        Zhanxuan Peng,
        Zihao Yuan,
        Ningbin Zhang,
        and Guoying Gu
    }
    
    \thanks{This work was supported by the National Natural Science Foundation of China under Grant 52025057 and Grant 52305029, and in part by the Science and Technology Commission of Shanghai Municipality under Grant 24511103400. \textit{(Corresponding author: Guoying Gu.)}}
    \thanks{Yueshi Dong, Jieji Ren, Zhenle Liu, Zhanxuan Peng, Zihao Yuan, Ningbin Zhang and Guoying Gu are with the State Key Laboratory of Mechanical System and Vibration, Shanghai Jiao Tong University, Shanghai, China, and also with the Robotics Institute, School of Mechanical Engineering, Shanghai Jiao Tong University, Shanghai, China (e-mail: dongys525@sjtu.edu.cn; jiejiren@sjtu.edu.cn; extraordinary@sjtu.edu.cn; steven2606@sjtu.edu.cn; yuanzihao@sjtu.edu.cn; zhangnb@sjtu.edu.cn; guguoying@sjtu.edu.cn).}
}

\maketitle

\begin{abstract}
Camera-based tactile sensors provide robots with a high-performance tactile sensing approach for environment perception and dexterous manipulation. However, achieving comprehensive environmental perception still requires cooperation with additional sensors, which makes the system bulky and limits its adaptability to unstructured environments. In this work, we present a vision-enhanced camera-based dual-modality sensor, which realizes full-scale distance sensing from 50 cm to -3 mm while simultaneously keeping ultra-high-resolution texture sensing and reconstruction capabilities.
Unlike conventional designs with fixed opaque gel layers, our sensor features a partially transparent sliding window, enabling mechanical switching between tactile and visual modes. 
For each sensing mode, a dynamic distance sensing model and a contact geometry reconstruction model are proposed.
Through integration with soft robotic fingers, we systematically evaluate the performance of each mode, as well as in their synergistic operation. Experimental results show robust distance tracking across various speeds, nanometer-scale roughness detection, and sub-millimeter 3D texture reconstruction. The combination of both modalities improves the robot’s efficiency in executing grasping tasks.
Furthermore, the embedded mechanical transmission in sensor allows for fine-grained intra-hand adjustments and precise manipulation, unlocking new capabilities for soft robotic hands.

\end{abstract}

\begin{IEEEkeywords}
Dual-modality sensor, tactile sensing, proximity sensing, intelligent grasping, fine-texture reconstruction, soft robotic hand, in-hand manipulation.
\end{IEEEkeywords}

\section{Introduction}

\IEEEPARstart{A}{mong} various exteroceptive modalities, tactile sensors play a crucial role in enabling robots to perceive contact-related information such as pressure, surface texture, and object geometry~\cite{doi:10.1089/soro.2020.0019},~\cite{doi:10.1126/scirobotics.adl0628},~\cite{xu2024dtactive}. Typically embedded in robotic fingertips or palms, these sensors allow robots to adapt to uncertainties in object shape, surface friction, and contact force during manipulation tasks. Over the past decade, a wide variety of tactile sensors have been developed based on diverse transduction mechanisms, including capacitive~\cite{2018Transparent},~\cite{yoo2018industrial}, piezoelectric~\cite{hammock201325th},~\cite{pi2014flexible} and fiber Bragg grating (FBG)~\cite{huang2019fiber}, etc. While these sensors offer high sensitivity and resolution, their practical deployment is often hindered by challenges such as complex wiring, signal crosstalk, and limited scalability, especially in soft or highly integrated robotic systems~\cite{zhang2025soft}~\cite{gu2023soft}~\cite{xu2024learning}. In response to these limitations, vision-based tactile sensing (VBTS) has emerged as a promising alternative~\cite{yuan2017gelsight}. By embedding a camera within an enclosed space and illuminating the contact interface with structured or directional lighting, VBTS enables the extraction of high-fidelity tactile information on a soft contact layer by interpreting visual changes in the contact surface. This approach allows simultaneous acquisition of surface geometry, contact force distribution, and fine textures~\cite{10746317},~\cite{taylor2021gelslim30highresolutionmeasurementshape},~\cite{2023Evetac}, making VBTS one of the most widely adopted tactile sensing methods in the field of dexterous manipulation.

While vision-based tactile sensing provides detailed contact information, a single modality is often insufficient for achieving comprehensive environmental perception in robotic operations. To overcome this limitation, robots commonly integrate data from peripheral sensors—such as proximity sensors and external cameras—to enhance spatial awareness and task adaptability. Among them, capacitive~\cite{9592677},~\cite{2024Proximity} proximity sensors are widely used for short-range detection and dynamic distance estimation. However, their sensing performance is highly susceptible to environmental factors such as the refractive index, ambient light, and the presence of surrounding objects, often leading to compromised accuracy, stability, and durability. Cameras, on the other hand, offer a more robust and versatile sensing modality by capturing rich visual information—including object color, contour, and depth—across a broader field of view. Nevertheless, visual occlusion remains a critical challenge~\cite{9592677}. To mitigate this, researchers have proposed multi-camera setups with optimized placement strategies~\cite{doi:10.1177/0278364919887447},~\cite{10577462},~\cite{10380224}. As a result, such large-scale perception systems often struggle to adapt to unstructured or dynamically changing environments, limiting their practical deployment in real-world robotic tasks. These challenges highlight the pressing need for a compact, unified sensing framework that can seamlessly switch between multi-modalities.

Therefore, several recent studies have attempted to enhance the full-scale perception capability of a single VBTS unit. These efforts aim to unify proximity and tactile sensing within a single hardware platform, thereby reducing system complexity. For instance, Shimonomura et al. additionally constructed an infrared total internal reflection environment within the transparent flexible contact layer, determining the measurement mode based on whether the infrared patterns could be extracted from the raw camera images, and realized adaptive grasping motion for a six degrees of freedom (DOF) robotic arm~\cite{7487126}. Wang et al. introduced a design with continuous UV lamp switching, identifying whether the sensor was in contact with the target by comparing the movement of marker points across each frame of the video stream~\cite{9812348}. However, both designs rely on passive mode switching triggered by contact, limiting their ability to perform proactive perception in complex environments. Hogan et al. trained a network to enable the measurement of 'approaching distance' for a specific target and effectively separated two modalities in continuous image frames captured by the camera~\cite{9832483}. While this design allows simultaneous perception of external object depth and internal contact force, it sacrifices detailed surface characterization due to optical complexity. As another representative work, Zhang et al. proposed TIRgel, a vision-based tactile sensor that leverages total internal reflection (TIR) within a transparent elastomer and achieves modality switching via focus adjustment~\cite{10224334}. However, it remains limited by the lack of explicit distance estimation capability and passive mode conversion based on optical parameters alone. The design of the dual-mode sensor that integrates long-distance proximity sensing and high-quality contact sensing has long remained a persistent challenge.

\begin{figure}[!t]
    \centering
    \includegraphics[width=\columnwidth,keepaspectratio]{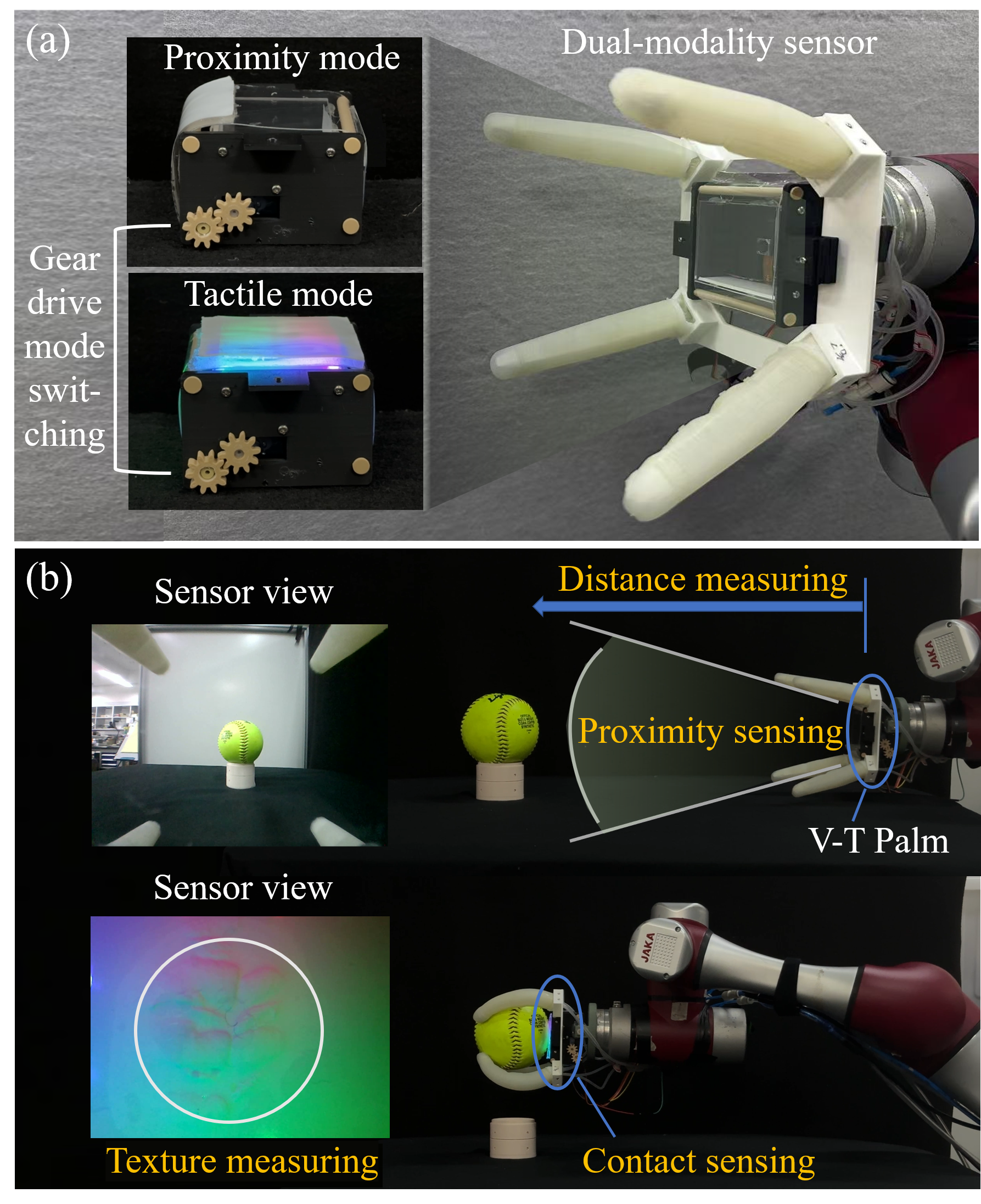}
    \caption{Illustration of two working modes of the V-T Palm and its application in a robotic system.}
    \label{fig:main}
\end{figure}

In this article, we propose a vision-enhanced proximity–tactile sensor that achieves active and unified dual-modality perception through a mechanically driven rotatable belt mechanism (see Fig.~\ref{fig:main}). By incorporating a partially opaque sliding window, the system can actively extend or retract an elastic opaque layer as needed. During tactile sensing mode, the elastic layer will be screwed out to serve as the outermost contact interface. By leveraging a photometric stereo algorithm~\cite{5206534}, the detailed contact surface geometry can be reconstructed. In the proximity sensing mode, the exposed zoom camera captures external visual data and enables robust long-range target tracking through the combination of monocular depth estimation~\cite{yang2024depthv2} and a lightweight segmentation pipeline, while avoiding the need for multi-camera calibration and complex occlusion processing.
By tightly integrating high-fidelity tactile reconstruction and long-range proximity perception into a single compact module, our design offers a scalable, efficient, and adaptable sensing solution for real-world robotic grasping and interaction tasks.

% ./imgs/2025.03.11-main graph-final.png

The experimental results show that the proposed sensing system can accurately measure the distance of the various targets under different speeds in real-time conditions. Furthermore, it can achieve a nanometer-scale roughness resolution and sub-millimeter texture reconstruction. By embedding into a soft gripper, the combination of both modalities improves the robot’s efficiency in executing grasping tasks. Meanwhile, due to the introduction of the transmission mechanism, the sensor itself has obtained an extra DOF of manipulation. Through an application experiment of card insertion, the possibility of precise adjustments within a soft gripper has been proofed. The contributions of this work are threefold:
\begin{itemize}
\item Modular sensing unit for robotic grasper: We introduce the dual-modality sensor as a palm (V-T Palm), a novel robotic palm prototype that seamlessly integrates proximity and tactile sensing to support the whole process of robotic grasping.
\item Extension of proximity sensing: achieve full-scale distance perception from 50 cm to -3 mm without interference with fine texture reconstruction of tactile sensing.
\item Fine-tuning operation ability: We coupling the sliding window to enable real-time sensing feedback and in-grasp pose adjustments of the soft robotic hand.
\end{itemize}

%%%%%%%%%%%%%%%%%%%%%%%%%%%%%%%%%%%%%%%%%%%%
%%%%%%%%%%%%%%%%%Design and fabrication%%%%%%%%%%%%%%%%%%%%
%%%%%%%%%%%%%%%%%%%%%%%%%%%%%%%%%%%%%%%%%%%%
\section{DESIGN AND FABRICATION}

\begin{figure*}[htbp]
    \centering
    \includegraphics[width=\textwidth,keepaspectratio]{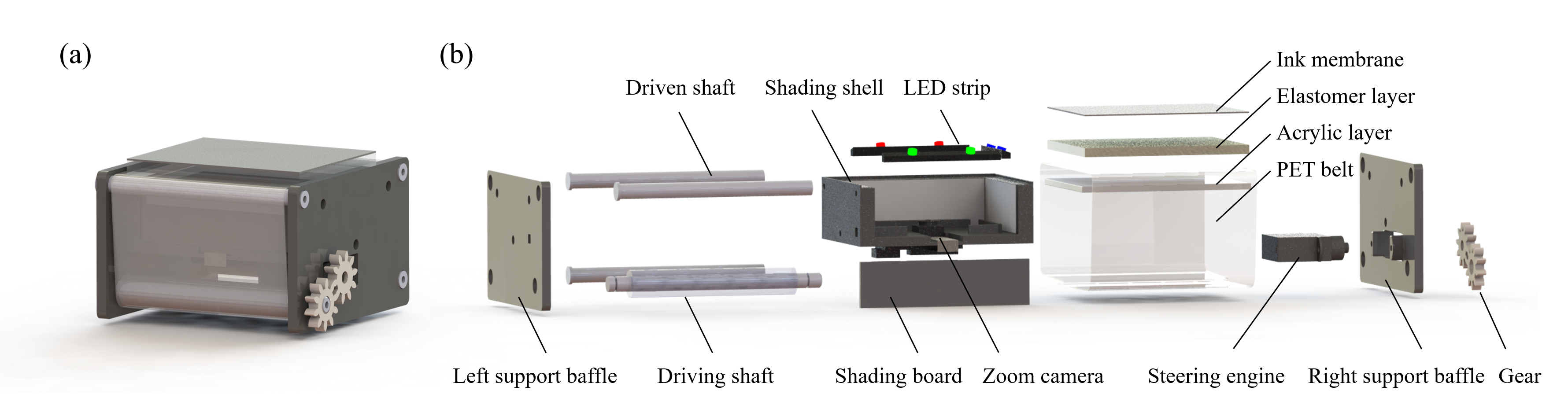}
    \caption{Design of V-T palm. (a) Transmission mechanism is introduced to integrate the sensing functions of two modes into a compact structure. (b) Detailed composition structure.}
    \label{fig:struc1}
\end{figure*}

\begin{figure}[!ht]
    \centering
      \includegraphics[width=\columnwidth,keepaspectratio]{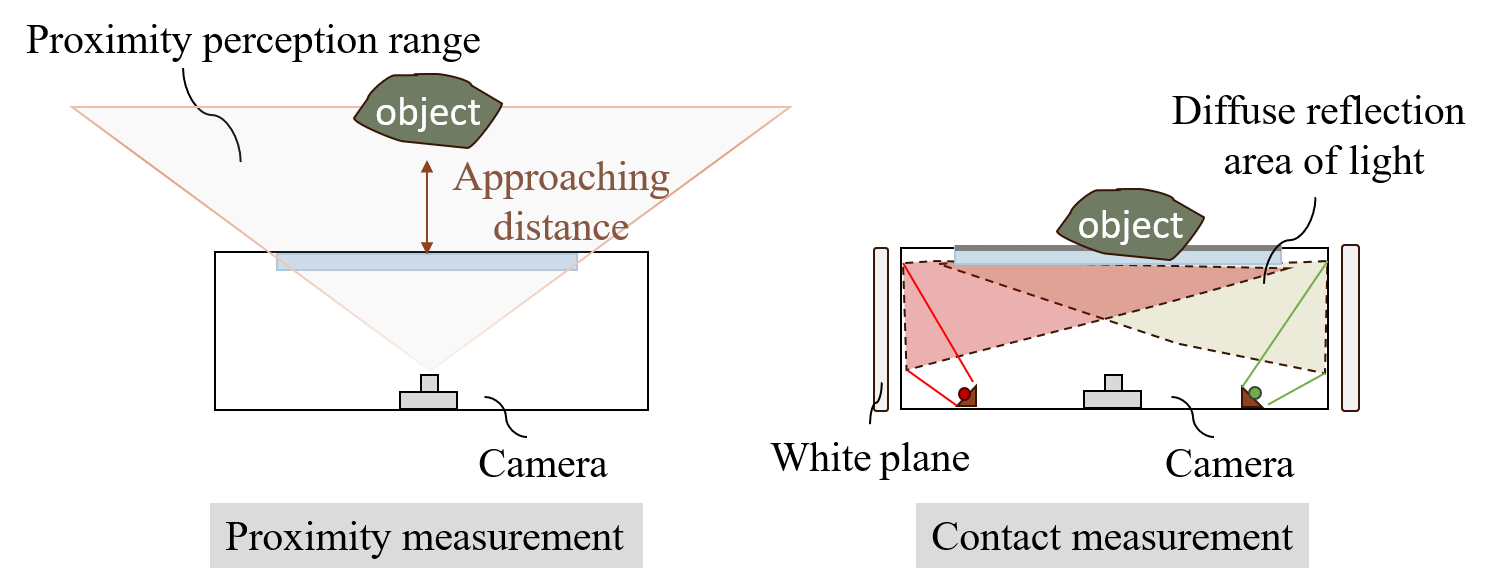}
    \caption{Measurement principle. The proximity sensing mode utilizes the camera to continuously capture images of external target objects. Under tactile mode, the LED strip is activated, enabling the acquisition of tactile information through the internal light field.}
    \label{fig:struc2}
\end{figure}

\begin{figure}[!ht]
    \centering
    \includegraphics[width=\columnwidth,keepaspectratio]{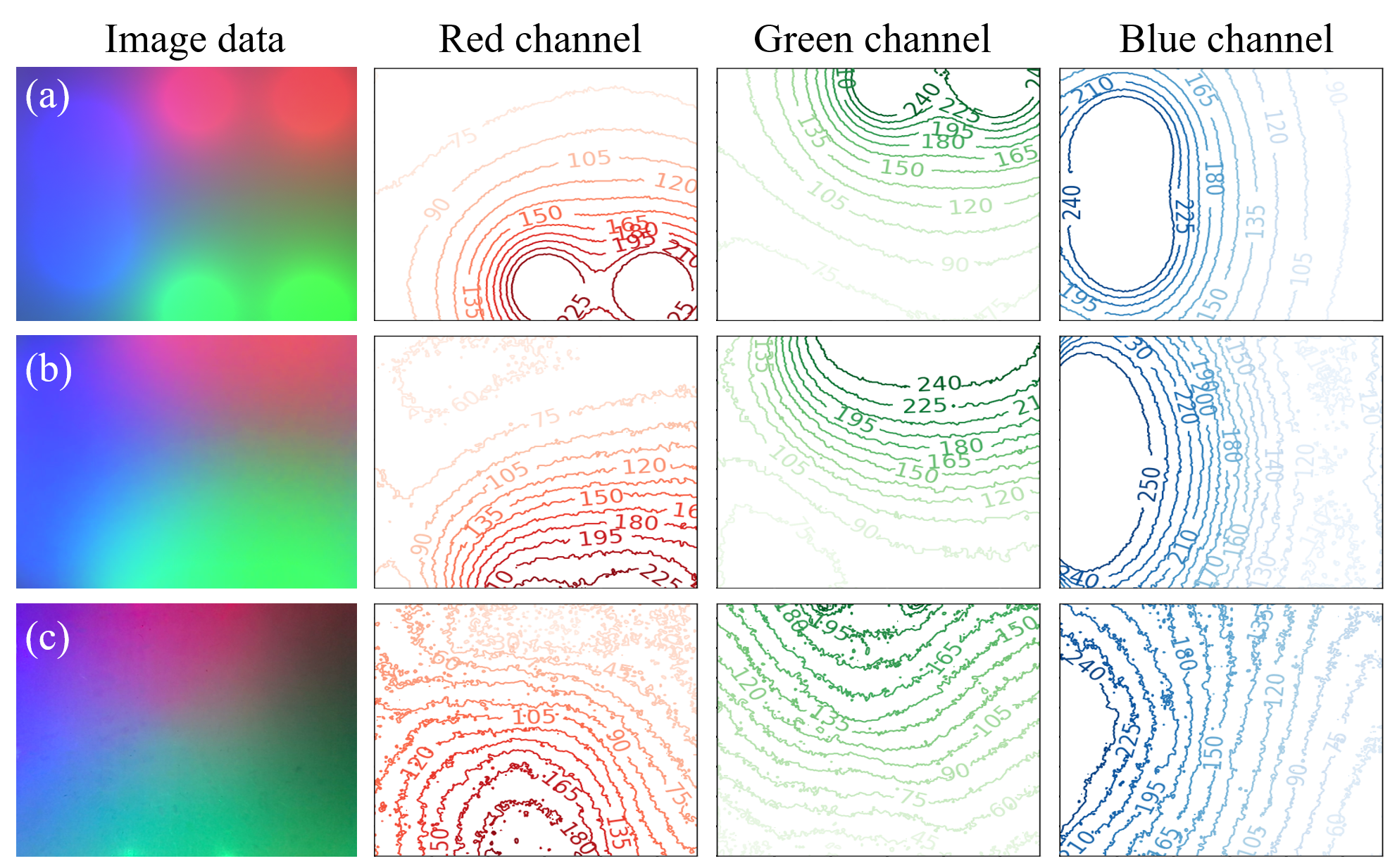}
    \caption{Comparison of different illumination methods. (a) and (b) show the comparison of light intensity distribution between the direct illumination scheme and the diffuse reflection scheme of lamp beads simulated in Blender. (c) illustrates the actual illumination condition of the sensor. }
    \label{fig:struc3}
\end{figure}
% fig4 光照设计效果展示-待替换.png

The overall design of the V-T Palm (see Fig.~\ref{fig:struc1}) can be broadly divided into two main components:  the modal switching module and the sensing module.
\subsection{Dual-mode Switching Module}

The rolling switching module is a key focus of this work. Although several effective rolling plane design approaches have been proposed~\cite{10161516},~\cite{10892188},~\cite{xu2024dtactive}, how to ensure that the transmission belt incorporates both completely opaque and fully transparent regions and preventing slippage during transmission poses a significant challenge to the design methodology. Pure PET has good optical transparency, and the light transmittance can usually reach 85\%-90\% in the visible light range (400-700 nm), which is close to the performance of glassy materials. At the same time, it has high hardness and relatively low surface energy, which can minimize friction with other internal parts and enable flexible rotation, making it an excellent choice for transmission baseband. We cast a 2 mm thick silicone layer on the outermost layer of the driving shaft. The adhesive friction, deformation friction, and intermolecular force generated by the characteristics of the silicone material ensure strong contact between the silicone layer and the PET belt and also achieve a stable friction transmission effect. The specific design details are as follows:

The entire transmission mechanism consists of two baffles, a driving roller, three passive rolling shafts, a conveyor belt, a $360^\circ$ steering gear, and two gears with a 1:1 transmission ratio. Most of these are fabricated using 3D printing (X1Carbon, Bambu Lab). The driving shaft is driven by the steering gear, and it coated with a silicone layer (PDMS 0030) using a single-stage casting process. The conveyor belt rotates without slipping due to the friction generated between the silicone ring and the PET belt. The whole mechanism is employed to drive the conveyor belt along the outermost ring of the sensor, enabling switching between proximity measurement and contact measurement. The conveyor belt is based on a rectangular PET strip measuring 180~mm~$\times$~55~mm. A flexible contact measurement module, composed of silica gel and reflective film, is adhered to the PET strip using silicone adhesive (Silpoxy, Ecoflex). Additionally, a hook-and-loop fastener is installed at the end of the strip, allowing for convenient re-tensioning of the conveyor belt as needed. When the distance between the target and the palm is detected as less than the threshold (e.g., 3 cm or 10 cm), a command is executed to rotate the elastic contact module above the camera, activating the tactile measurement mode. Conversely, when the module is retracted, the high transparency of the PET material ensures that the camera's ability to capture the external environment remains unaffected, enabling the proximity measurement mode. The operating principles of the two sensing modes are illustrated in Fig.~\ref{fig:struc2}. The power supply control of the LED lamp and the steering gear’s control signals are coordinated and transmitted by an Arduino board(Arduino Mega 2560, Arduino SRL), following the output commands from the algorithm running on the host computer.

\subsection{Sensor Module}
Both measurement modes share the same design foundation. A monocular zoom camera (OV5640) with an effective viewing angle of 120 degrees is installed inside the sensor, capturing real-time RGB images at a resolution of 5 megapixels and a frame rate of 30 frames per
second (fps). The zoom lens features both active and passive focusing capabilities, enabling clear imaging of target objects at varying distances from the camera. On the top of the sensor is a transparent acrylic plate with a thickness of 3~mm, which provides a reliable support strength of 0\textendash 200~\text{kPa}
 for the flexible contact measurement layer of the sensor in addition to providing a light refractive index close to air and PET material, provides a reliable support strength for the stable grasping task.
 
The inner cavity of the sensor is 3D printed by black Polylactic Acid (PLA) material, which provides a stable inner space for the optical coding of tactile information. A strip structure embedded with RGB LED strip is designed at the lower edge of the bottom near the three side walls so that the plane normal vector of the lamp strip directly faces the three sides.
Referring to the principle of diffuse reflection, we applied an even coating of white paint to the three walls. This design enables the red, green, and blue LED point light sources to transform into surface light sources through reflection, thereby reducing the influence of concentrated beam refraction between different light-transmitting materials and creating a more uniform lighting conditions for tactile imaging. Using Blender simulations, we compared our method with the lighting effects produced by direct point light source illumination, as shown in Fig.~\ref{fig:struc3}. The results demonstrate that the light intensity distribution across the lens is significantly more uniform under the proposed illuminated solution, the effective sensing area is also further expanded, which enhances the quality of raw sensing image data.

%%%%%%%%%%%%%%%%%%%%%%%%%%%%%%%%%%%%%%%%%%%%
%%%%%%%%%%%%%%%%%Method%%%%%%%%%%%%%%%%%%%%
%%%%%%%%%%%%%%%%%%%%%%%%%%%%%%%%%%%%%%%%%%%%
\section{METHOD}

% \subsection{Calibration and Validation}
To achieve high-quality object perception in both sensing states, we developed a distance measurement framework and a contact texture reconstruction approach. Specifically, for proximity sensing, we formulated a nonlinear mapping between real-world and camera coordinates, tailored to object approach scenarios. In tactile sensing, we established a correlation between the geometric characteristics of passive surface deformations and the internal light field, enabling high-resolution tactile perception.

\subsection{Nonlinear Dynamic Mapping Model of Proximity Sensing}

\begin{figure}[!t]
    \centering
    \includegraphics[width=\columnwidth,keepaspectratio]{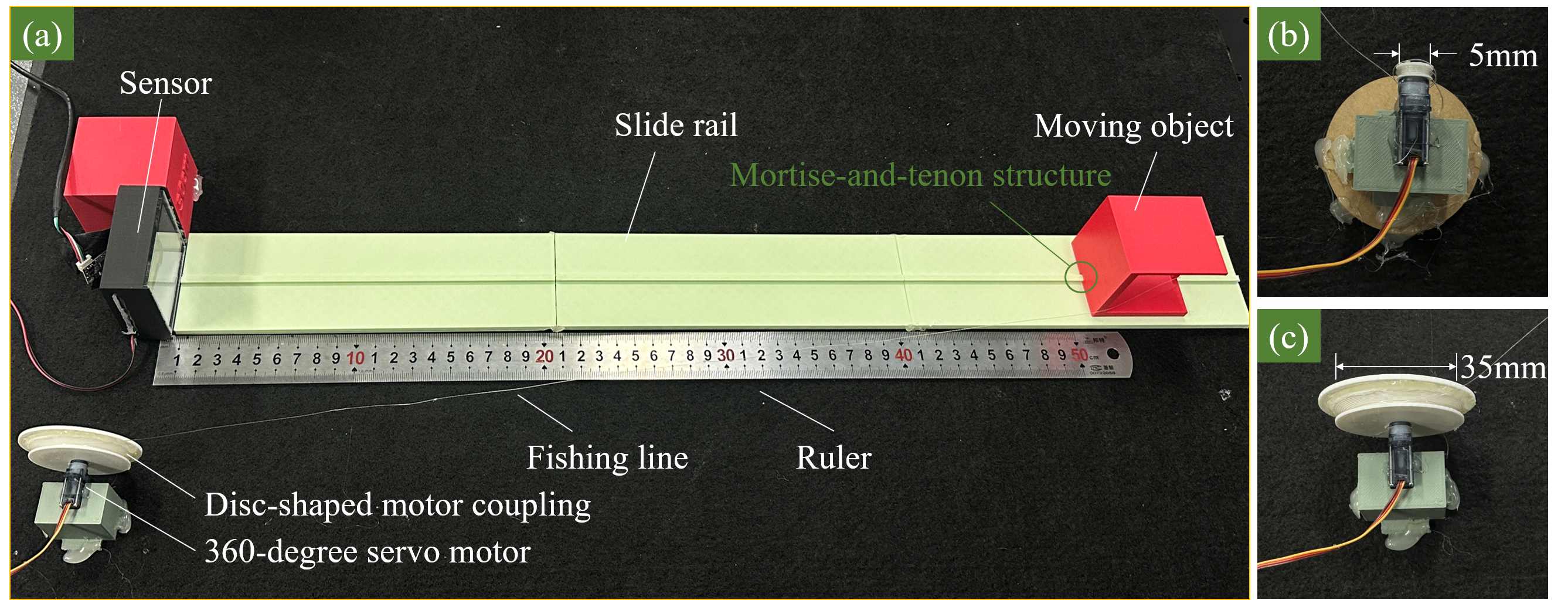} 
    \caption{ {Experimental platform setup of proximity sensing. (a) illustrates the detailed composition of the test platform; (b) and (c) show two kinds of disc steering gear couplings with inner diameters of 5 mm and 35 mm, which realize the movement of the target object at different speeds}.}
    \label{fig:chara2}
\end{figure}

In our system, the internal parameters of the camera change dynamically in real time as the target moves, rendering the simple pinhole camera model~\cite{andrew2001multiple} no longer applicable. To address this issue, an experiment is devised to recalibrate the ranging system, as illustrated in Fig.~\ref{fig:chara2}. We let a block move along the central axis of the camera-captured area from far to near within a distance of $D\in[0,50] \, \mathrm{cm}$ under randomly various velocities. The detailed setup of the platform will be illustrated in Section IV.

\begin{figure*}[htbp]
    \centering
    \includegraphics[width=\textwidth,keepaspectratio]{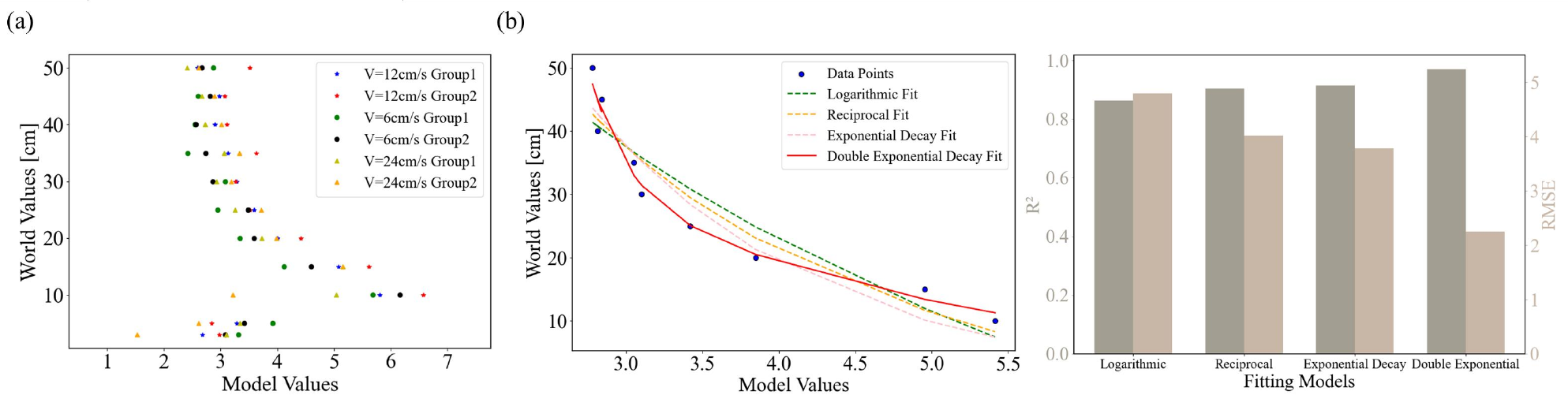}
    \caption{ {Development of a nonlinear dynamic mapping model for proximity sensing modality. (a) illustrates the distribution of ($Z_{img}$, $Z_{world}$) across different experimental groups, all of which exhibit a similar distribution pattern. (b) compares the fitting performance of different monotonic decreasing models on the experimental data. Among them, the Double Exponential Decay model exhibits the best fitting accuracy, with $R^2 = 0.9697$ and $\mathrm{RMSE} = 2.2462$.}}
    \label{fig:append1}
\end{figure*}

For each group of data collection, we extract the video frames corresponding to the target world distance of $D=\{3,5,10,15,20,25,30,35,40,45,50\} \mathrm{cm}$, calculate the predicted distance of the target with a combination of U-NET segmentation network trained on our custom dataset and a pre-trained monocular depth estimation algorithm~\cite{yang2024depthv2}. By observing the distribution law of data (Fig.~\ref{fig:append1}(a)), we found that the data generally presents an obvious two-stage distribution trend.  Especially for the interval of $D\in [10,50] \, \mathrm{cm}$, the predicted value presents a monotonic increasing trend with the approach of the distance of the target object. We employed several inverse proportional models to fit the data and ultimately identified the most appropriate expression to characterize the underlying data trends (Fig.~\ref{fig:append1}(b)). Among these models, the double exponential decay model provides the best fit:

\begin{equation}
    y = a \cdot e^{-b \cdot x} + c\cdot e^{-d \cdot x}
\end{equation}
where:
    \begin{itemize}
        \item $y$ : represents the distance value $\ Z_{world}$ in the real-world coordinate system.
        \item $x$: represents the measured depth value $\ Z_{img}$ in the camera coordinate system, which is equal to the mean value of image target segmentation $\mu_{\Omega}$.
    \end{itemize}

\begin{equation}
\begin{aligned}
    \mu_{\Omega} &= \frac{\sum_{(u, v) \in \Omega} D(u, v)}{\lvert \Omega \rvert}\\
    % \ Z_{img} &= \mu_D\\
    % \ Z_{world} &= \frac{43.1319}{Z_{img}} - 2.3122\\
\end{aligned}
\end{equation}
where:
\begin{itemize}
    \item $\Omega = \{(u, v) \mid M(u, v) \neq 0\}$: represents the coordinate set of pixels in the non-zero region.
    \item \(D(u, v)\): represents the whole depth map predicted by the algorithm.
    % \item \(\mathbf{K}\) is the camera intrinsic matrix.
\end{itemize}

From the obtained fitting results, the parameters were determined as follows: 
$a = 85.9058$, $b = 0.3754$, $c = 1.5110 \times 10^6$, and $d = 4.0941$, where 
$R^2 = 0.9697$ and $\mathrm{RMSE} = 2.2462$.

\subsection{Geometry Reconstruction of Tactile Sensing}

Under tactile mode, we set the height equation of the touching surface as \( z = f(u, v) \), then the surface normal can be express as follows:
\begin{equation}
\begin{aligned}
% \[
\mathbf{N}(u, v) = \left( \frac{\partial f}{\partial u}, \frac{\partial f}{\partial v}, -1 \right)
% \]
\end{aligned}
\end{equation}
% N(x,y) = (pf/px,pf/py,-1)

According to the principle of photometric stereo algorithm ~\cite{yuan2017gelsight}, we add the local pixels position of lighting into mapping consideration to establish a local mapping relationship for each pixel by constructing a neural network, correlating the intensity values of the R, G, and B channels with the gradients along the X and Y axes, as represented by the following equation:

\begin{equation}
\begin{aligned}
% \[
% R_i(I_R, I_G, I_B, x, y) = (\frac{\partial f}{\partial x}, \frac{\partial f}{\partial y})
R_i(I_R, I_G, I_B, u, v) = (G_u, G_v)
% \]
\label{eq:mapping}
\end{aligned}
\end{equation}
where:
\begin{itemize}
    \item $R_i$: represents the local pixel's mapping relationship between light intensity and geometry gradients.
    % \item \((x,y)\): indicates the spatial coordinates of the pixel.
    \item  \((I_R, I_G, I_B)\): denotes the light intensity values of the red, green, and blue channels, respectively.
    % \item \((x,y)\): indicates the spatial coordinates of the pixel.
    \item \((G_u, G_v)\) \(=\) \(\left(\frac{\partial f}{\partial u}, \frac{\partial f}{\partial v}\right)\).
    % \item \((G_u, G_v)\) = \nabla z .

\end{itemize}

To obtain the training dataset, we randomly pressed a standard ball with a diameter $r$ = 5mm on the sensing surface, assuming the radius of intersecting circle pattern is $r*$. Based on the geometric relationship, the distance $h$ between the center of the sphere and the plane can be determined as:

\begin{equation}
\begin{aligned}
        h = \sqrt{r^2 - r^{*2}}
\end{aligned}
\end{equation}

For the intersecting portion of the spherical surface, the equation of the sphere can be expressed as:

\begin{equation}
\begin{aligned}
    z = h - \sqrt{r^2 - u^2 - v^2}
\end{aligned}
\end{equation}

Take the partial derivatives of u and v respectively, we can get the geometric gradient expression of each point in the intersecting spherical part:

\begin{subequations}
    \begin{equation}
    \nabla z = 
    \left(
    \frac{u}{\sqrt{r^2 - u^2 - v^2}}, 
    \frac{v}{\sqrt{r^2 - u^2 - v^2}}
    \right)
    \end{equation}
    \begin{equation}
    (G_u, G_v) = \nabla z.
    \end{equation}
\end{subequations}
Where $(u, v)$ satisfies the constraint conditions of intersecting circles:

\begin{equation}
    u^2 + v^2 \leq r^{*2}
\end{equation}

% We acquired 30 images with a resolution of $1280\times960$ pixels.
A total of 30 images with a resolution of $1280\times960$ pixels were collected. After data cleaning, 842,400 data entries were obtained. Then $R_i$ for each pixel position can be obtained by the neural network built above.

Therefore, the gradient map $\vec{g}$ for each test image will be calculated, then the depth map $\phi $ of the contact surface can be reconstructed by solving a two-dimensional Poisson equation.

\begin{equation}
\begin{aligned}
    \nabla^2 \phi = \nabla \cdot \vec{g}
\end{aligned}
\end{equation}
Where $\nabla^2 \phi$ is the Laplacian of the depth map, and $\nabla \cdot \vec{g}$ represents the divergence of the gradient map.

Given the high computational speed requirements for image reconstruction in the target application, the Fourier transform method was selected to solve the Poisson equation. This method is well-suited for global solutions and demonstrates greater robustness when applied to data with conventional image characteristics. The computational procedure is as follows:

Firstly, perform fast Fourier transform operation on $\nabla \cdot \vec{g}$,
\begin{equation}
   F = \mathcal{F}(\nabla \cdot \vec{g}),
% \begin{aligned}
%     \nabla^2 \phi = \nabla \cdot \vec{g}
% \end{aligned}
\end{equation}

Secondly, solve the Poisson equation based on frequency domain division,

\begin{equation}
    \mathcal{F}(\phi ) = \frac{\mathcal{F}(\nabla \cdot \vec{g})}{-k_x^2 - k_y^2},
\end{equation}
where \( k_x \) and \( k_y \) are frequency variables.

Finally, perform the inverse Fourier transform on the solution, then the depth map $\phi$ can be reconstructed.,

\begin{equation}
    \phi = \mathcal{F}^{-1}(\mathcal{F}(\phi)).
\end{equation}

\subsection{Switching Mechanism Between Two Sensing Modes}
\begin{figure*}[htbp]
    \centering 
    \includegraphics[width=\textwidth,keepaspectratio]{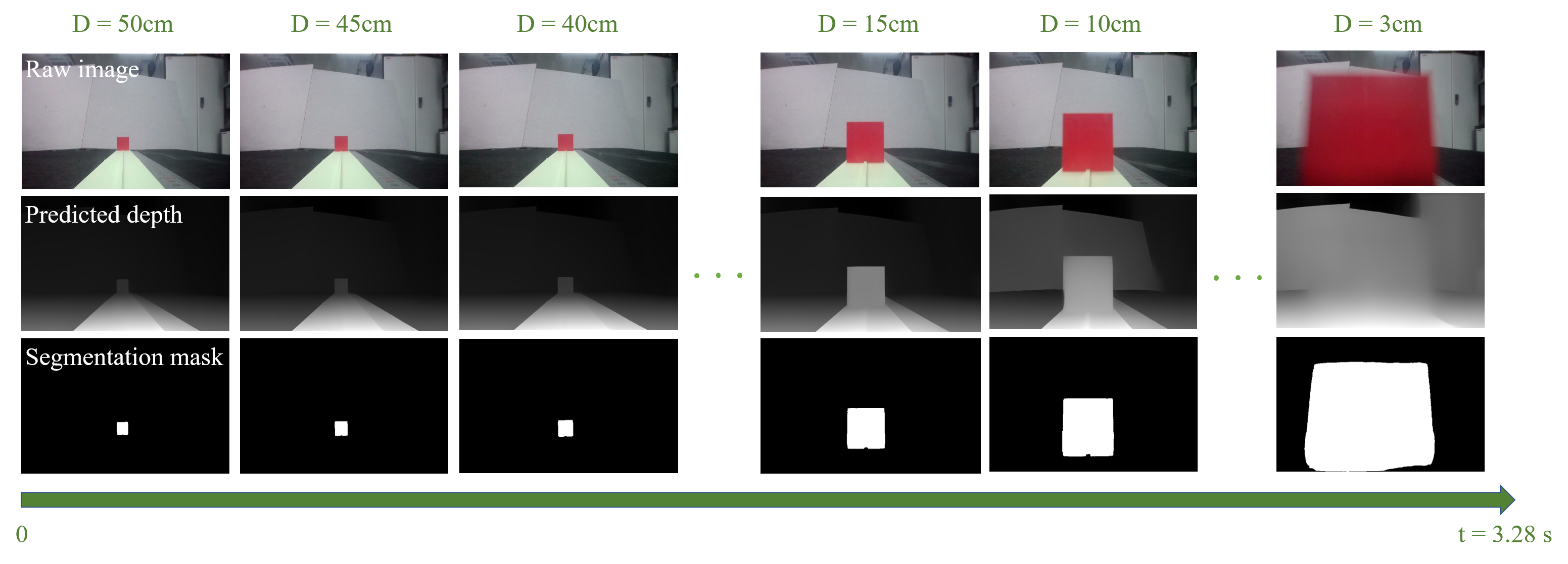} % Fig 8 save command: plt.savefig(save_path, dpi=300, bbox_inches='tight', format='png')
    \caption{ {Schematic of the distance measurement algorithm. Using V = 12.5 cm/s as an example, the original image, depth map, and target segmentation results of each frame are sequentially displayed from top to bottom as time progresses.}}
    \label{fig:chara3}
\end{figure*}
\begin{figure*}[htbp]
    \centering 
    \includegraphics[width=\textwidth,keepaspectratio]{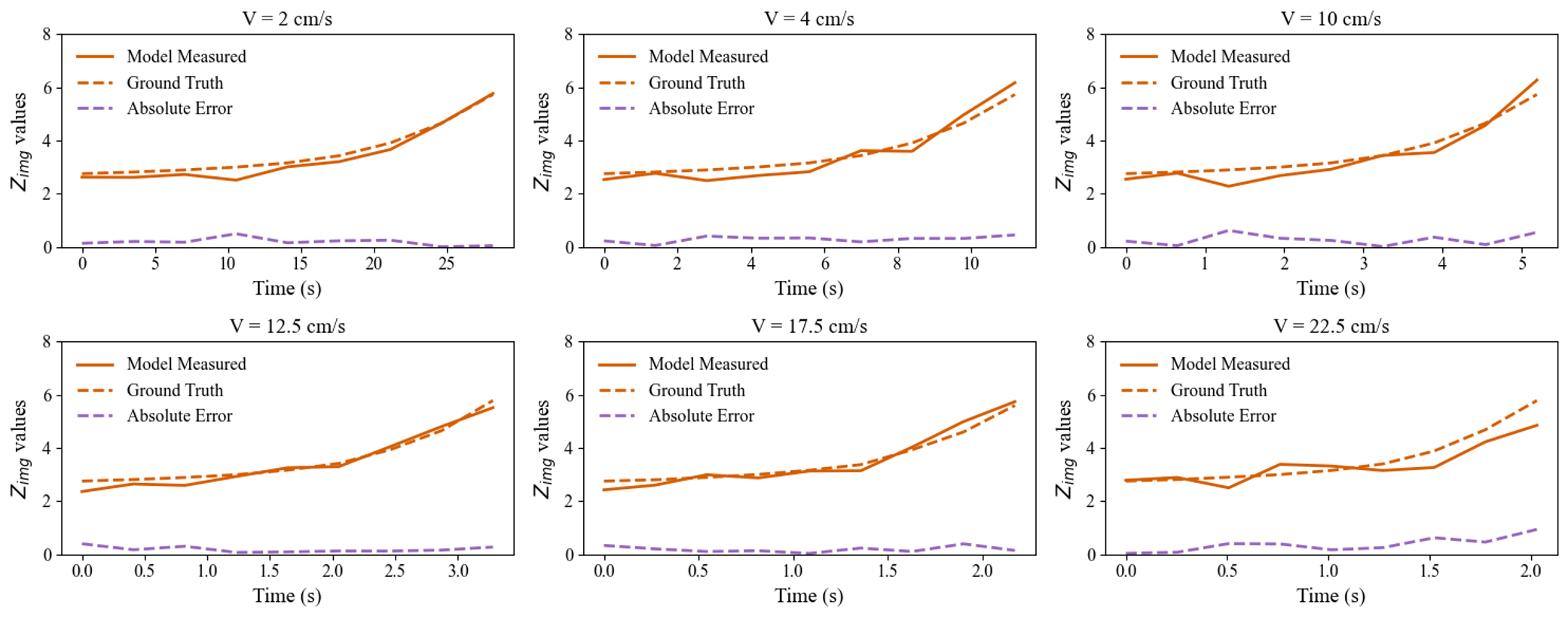} % Fig 8 save command: plt.savefig(save_path, dpi=300, bbox_inches='tight', format='png')
    \caption{ {Experimental results of the distance measurement tests under different speeds with the nonlinear dynamic mapping model of proximity sensing.}}
    \label{fig:chara4}
\end{figure*}
We leverage the friction between the silicone film and the PET belt to achieve mode switching, ensuring the stable movement of the flexible contact film (see Supplementary Video). A continuous rotation servo motor with a 360-degree range is precisely controlled using pulse-width modulation (PWM) signals via an Arduino-based embedded system. The motor operates at predefined speeds and directions based on serial commands. The control algorithm dynamically adjusts the pulse width of the control signal to achieve the desired rotational direction and speed.

Through systematic experiments, we identified the pulse width of 100~\textmu s to initiate counterclockwise rotation, while 2500~\textmu s to trigger clockwise motion, at which the transmission efficiency between the drive belt and the rotating shaft is maximized, with negligible slippage. When the target approaches the predefined distance threshold $D_t$ = 10 cm, the mode-switching mechanism is activated, triggering the rotation of the servo motor and the illumination of the LED strip, thereby transitioning to the tactile measurement mode. To ensure precise control, the motor’s rotation duration is fine-tuned using timed delays, allowing it to complete predefined movement before stopping.

%%%%%%%%%%%%%%%%%%%%%%%%%%%%%%%%%%%%%%%%%%%%
%%%%%%%%%%%%%%%%%Performance Characterization%%%%%%%%%%%%%%%%%%%%
%%%%%%%%%%%%%%%%%%%%%%%%%%%%%%%%%%%%%%%%%%%%

\section{Performance Characterization}

\subsection{Distance Measurement Accuracy of Proximity Perception} 

We let the target block of the proximity ranging experimental platform move at a series of set speeds ($V_i$ $\in$ \{2, 4, 10, 12.5, 17.5, 22.5\} $\mathrm{cm/s}$) to measure the reliability of the algorithm. The experimental set-up is detailed in Fig.~\ref{fig:chara2}. We fabricated the test block and slide rail by 3D printing technology (X1Carbon, Bambu Lab). To ensure controlled movement of the block in the desired direction, we designed the connection between the block and the slide rail as a mortise-and-tenon structure. The object block itself was designed with a "concave shape" to appear as a standard square from the perspective of the sensor. Inside the cavity of the block, a circular ring with an inner diameter of 2 mm and an outer diameter of 4 mm was incorporated. A transparent fishing line with a diameter of 1 mm was fixed to this ring, with its other end secured at the far end of the experimental platform. A 360-degree servo motor was placed at the far end of the platform, with its rotation speed controlled via a PWM signal emitted by an Arduino mega 2560 board. To achieve the desired retraction speed of the fishing line, we designed disc-shaped servo motor couplings of different sizes. The appropriate disc size for each experiment was determined through tests and detailed calculation. Additionally, a 2-mm-deep groove was created along the side of the disc to collect the retracted fishing line.

The sensor captured the block approaching video at all velocities, with a frame rate of 30 fps and a resolution of $1280 \times 960$ pixels per frame. 9 images are extracted from each video, in which the real-world distance of object should close to $D_i \in$ \{50, 45, 40, 35, 30, 25, 20, 15, 10\} \text{cm}. As shown in Fig.~\ref{fig:chara3}, each original image is processed by the pre-trained DepthAnythingV2 model, generating a complete depth map. Simultaneously, the target segmentation model, trained on a self-constructed dataset using the U-Net architecture, is applied to the original image, producing a mask map containing the position information of the target object. The predicted result \( Z_{img} \) for each image is obtained by multiplying the mask with the depth map and averaging the result. For each velocity, the experiment was repeated twice.

As shown in Fig.~\ref{fig:chara4}, the average absolute error of the measured values is of the order of \( 10^{-1} \), which indicates that the distance ranging algorithm performs well at the normal moving speed of the operating scene. When the target speed is at 17.5 cm/s, the absolute error is minimized, reaching approximately 0.183.

\subsection{Fine Texture Capture and Morphological Reconstruction}

\begin{figure}[!t]
    \centering
    \includegraphics[width=\columnwidth,keepaspectratio]{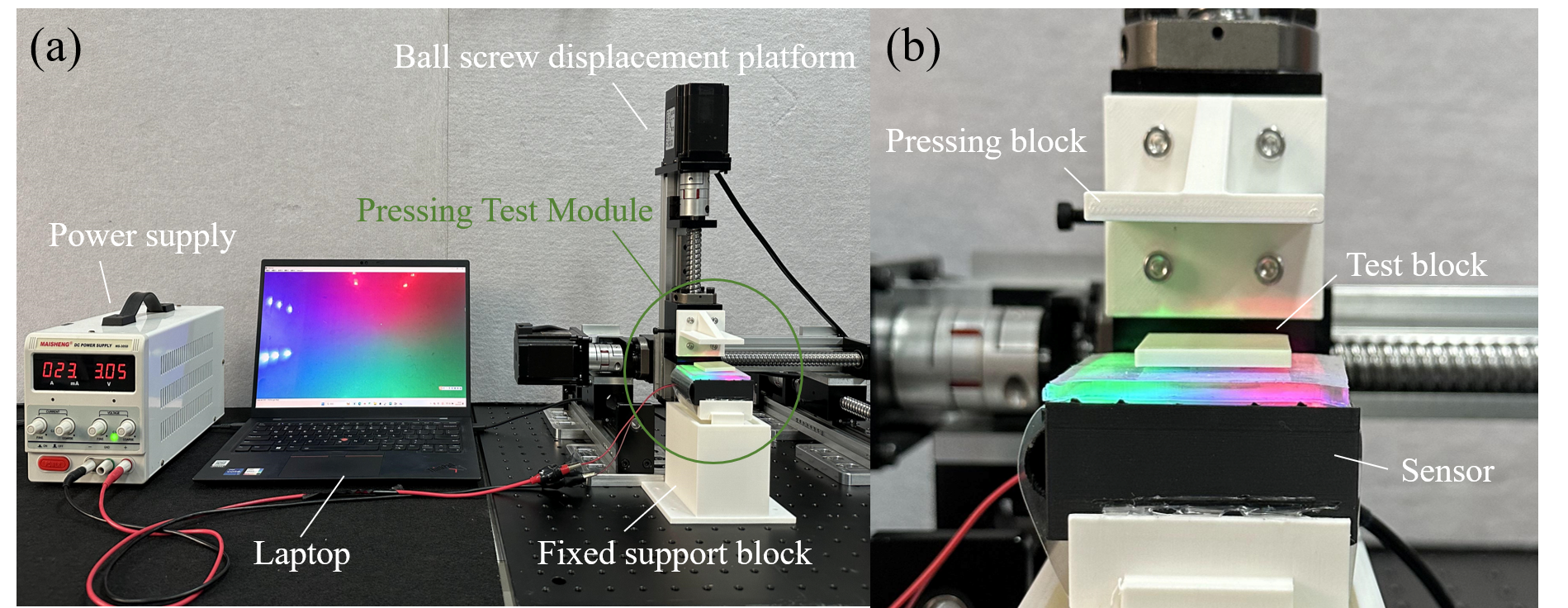}% Fig6_times_new_roman.png
    \caption{Experimental platform for tactile sensing. (a) illustrates the overall layout of the tactile perception experimental platform; (b) ensures the resultant normal force remains perpendicular to the sensing plane to minimize measurement errors.}
    \label{fig:chara_partb_1}
\end{figure}

Two characterization experiments are conducted to evaluate the sensor's tactile sensing ability, referencing the tactile sensing function of human skin. The first key metric assesses the sensor's detection threshold for surface roughness. The second metric evaluates the effectiveness of contact surface geometry reconstruction.
The experimental test platform is set up as shown in Fig.~\ref{fig:chara_partb_1}. The tactile sensing unit of the V-T Palm is mounted on a 3D-printed immovable support platform. Within the sensor, the RGB LED strip is powered by a DC power supply (Maisheng, China). A laptop is used to acquire high-resolution tactile images from the sensor’s built-in OV5640 camera modules. A 3D-printed planar pressing plate is mounted on a three-axis ball screw displacement platform, moving vertically along the screw slider, perpendicular to the horizontal plane. This design ensures that the resultant normal force applied to all measured objects remains consistently perpendicular to the sensing plane, thereby minimizing measurement errors. All original data for the tactile modality characterization experiment were acquired using the experimental platform developed above.

\subsubsection{Identification of Contact Surface Roughness}
In this experiment, sandpaper samples with varying roughness (grit sizes of 150, 280 and 500 mesh) were sequentially applied to the tactile sensor. The sensor's ability to perceive fine textures was assessed by analyzing the texture features extracted from the acquired data images. 

\begin{figure}[!t]
    \centering
    \includegraphics[width=\columnwidth,keepaspectratio]{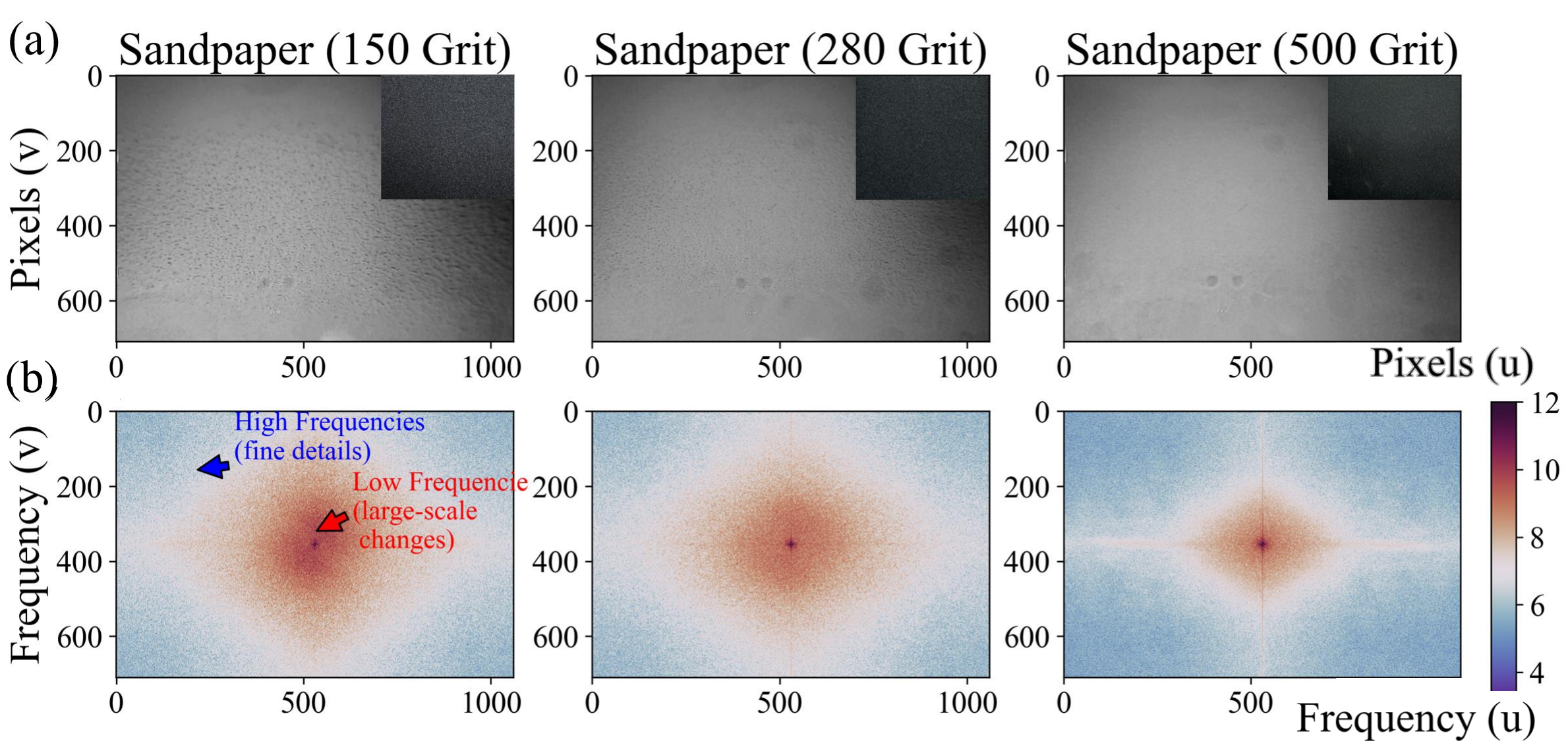}
    \caption{Frequency domain analysis of tactile images. (a) presents the grayscale-processed tactile data of sandpapers with different grit sizes, with the real photo of the corresponding sandpaper shown in the upper right corner. (b) illustrates the frequency amplitude spectrum visualized on a logarithmic scale.}
    \label{fig:chara_partb_2}
\end{figure}

In this process, two kinds of tactile data are collected separately, one is a tactile image without any pressing trace, the other is a tactile image obtained by pressing sandpaper with different roughness on the measurement surface. The difference images were obtained by subtracting the tactile images without contact from those acquired under sandpaper compression with varying roughness. A logarithmic scale was employed to improve the visibility of the amplitude spectrum, followed by zero-frequency shifting to the center of the spectrum. Fig.~\ref{fig:chara_partb_2}.(a) presents the grayscale representation of the original tactile image, while Fig.~\ref{fig:chara_partb_2}.(b) illustrates the processed results obtained using the proposed method. As shown in the illustrated results, the red region at the center corresponds to the low-frequency components, which represent large-scale texture features, whereas the blue regions at the image periphery indicate fine texture details. As the grit size of the sandpaper increases, the area of the low-frequency region in the analysis results decreases significantly, while the high-frequency region expands correspondingly. This demonstrates the sensor’s capability to effectively distinguish surface roughness among sandpapers with different grit sizes.

% Fig.10_sandpaper_analyse_1.png
The human tactile perception system is most sensitive to fine textures with a resolution of approximately 0.5–1 mm, which corresponds to sandpaper in the 60–80 mesh range. The experimental results demonstrate that the proposed sensor achieves a fine texture resolution capability approximately 6 to 8 times higher than that of the human tactile perception system.

\subsubsection{Reconstruction of Contact Surface Geometry}
\begin{figure*}[htbp]
    \centering
    \includegraphics[width=\textwidth,keepaspectratio]{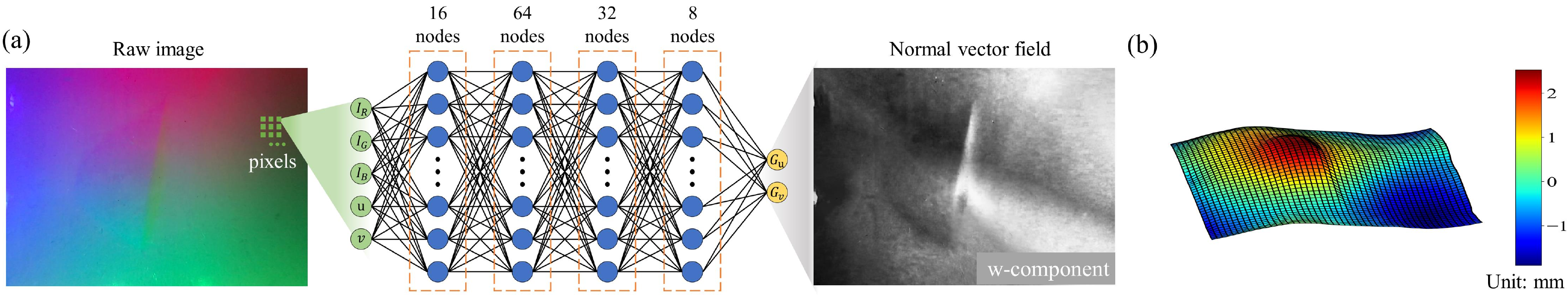}
    \caption{Schematic of the tactile data surface reconstruction process. (a) detailed illustrates the network architecture: A mapping from local pixel light intensities to geometric gradients. (b) depicts the surface geometry obtained through Poisson reconstruction.}
    \label{fig:chara_partb_3}
\end{figure*}

% \textcolor{blue}{write for 20250207, from this part , note that there are some data for platform establishment}
In this section, the geometry of the contact sensing surface is reconstructed using the previously introduced photometric stereo method. As shown in Fig.~\ref{fig:chara_partb_3},  we implement a fully connected neural network (FCNN) in PyTorch for tactile gradient image computation. The network consists of an input layer, four hidden layers, and an output layer. The input layer receives five-dimensional feature vectors (as shown in Eq.~(\ref{eq:mapping})), and each hidden layer comprises a fully connected (FC) layer followed by a ReLU activation function. The numbers of neurons in each hidden layer are 16, 64, 32, and 8, respectively. A dropout layer with a probability of 0.3 is applied before the final output layer $(G_u, G_v)$ to mitigate overfitting. After obtaining the gradient map in $(u, v)$ directions, the complete three-dimensional morphology of the tactile measurement surface can be obtained through Poisson reconstruction.

\begin{figure}[!t]
    \centering
    \includegraphics[width=\columnwidth,keepaspectratio]{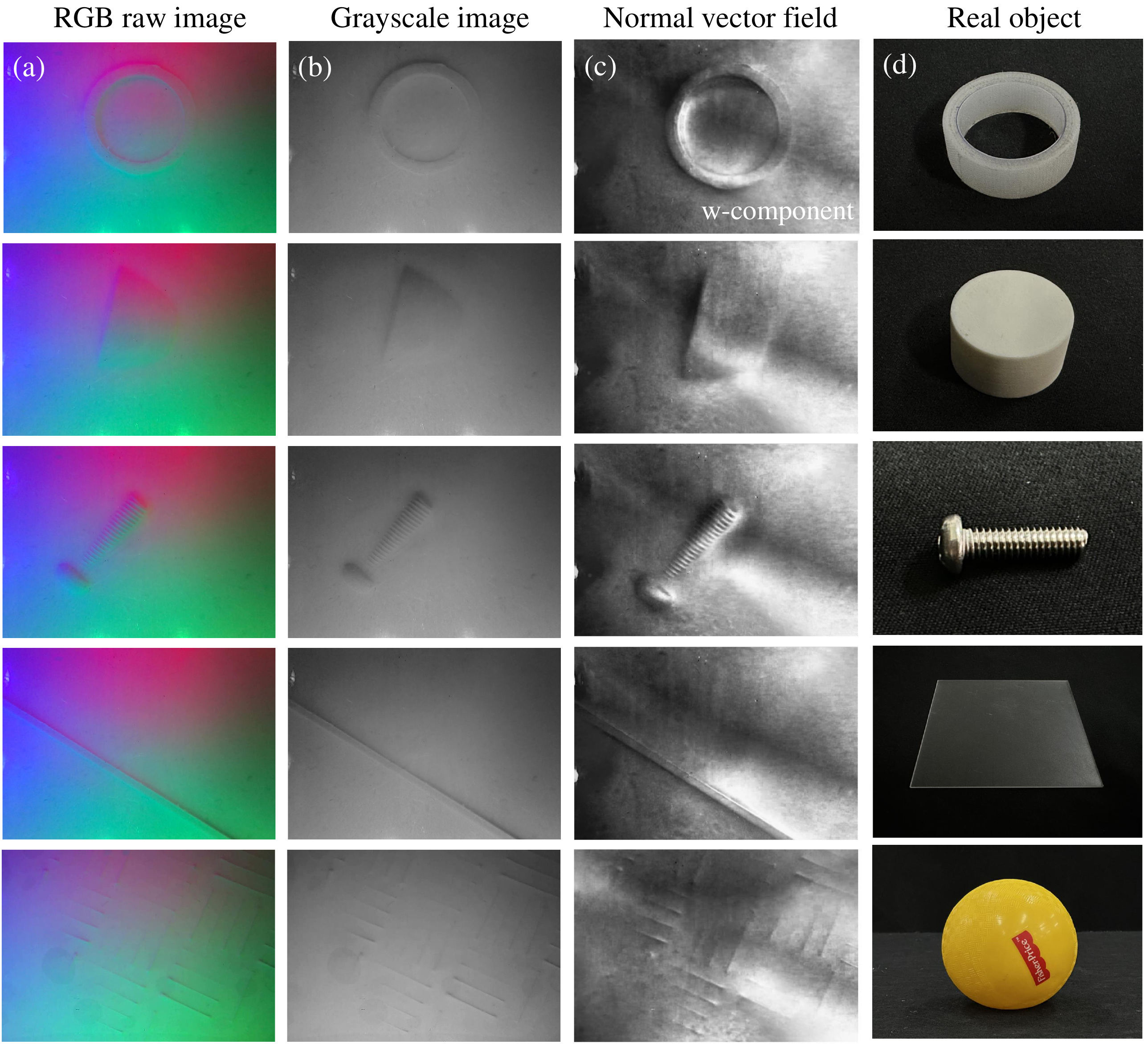}% 
    % FIG.11_final2_20250228_tactile_texture_reconstruction_times_new_roman.png
    % Fig6_times_new_roman.png
      \caption{Reconstruction performance of geometric texture. (a) shows the raw image data captured by the palm sensor. (b) displays the grayscale images of raw data, illustrating geometric information before processing. (c) shows the normal surface in w-axis calculated by the proposed FCNN algorithm, which indicates the depth information and surface orientation, and (d) presents the real objects.}
    \label{fig:chara_partb_4}
\end{figure}
The model is trained using the Adam optimizer with a learning rate of 3$\times$10$^{-5}$ to minimize the L1 loss function. To enhance convergence stability, an early stop mechanism is implemented, terminating the training process when the loss does not decrease for ten consecutive epochs. Furthermore, the training process is performed for 120 epochs, resulting in a regression model with a stable test mean squared error (MSE) of approximately 0.04.

Several objects commonly manipulated by robotic grippers were selected to evaluate the texture reconstruction performance of the V-T Palm, with the results presented in Fig.~\ref{fig:chara_partb_4}. To better visualize the depth information along the contact surface, the component perpendicular to the image coordinate plane was extracted and plotted using the Matplotlib library. Compared to the basic image processing method (Fig.~\ref{fig:chara_partb_4}(b)), the reconstructed surface not only preserves critical contact information but also enhances the texture details of the whole surface. This approach meets the requirements for surface texture extraction following typical grasping tasks performed by robotic grippers.

%%%%%%%%%%%%%%%%%%%%%%%%%%%%%%%%%%%%%%%%%%%%
%%%%%%%%%%%%%%%%%Experiment-APPLICATIONS%%%%%%%%%%%%%%%%%%%%
%%%%%%%%%%%%%%%%%%%%%%%%%%%%%%%%%%%%%%%%%%%%

\section{APPLICATIONS}
\subsection{Pre-plan Grasping and Subtlety Identification}

Grasping and identification of target objects is a common task in industrial applications. However, developing robust pre-grasp strategies, and identifying those with subtle texture differences (similar appearance, including color and shape) presents a challenge for the design and control of the entire mechanical system. 
Therefore, in this work, we integrate the palm and finger actuators together and investigate the comprehensive perceptual advantages of our sensing approach over individual sensing modalities (proximity sensing, external vision, and tactile sensing) in unstructured approach-to-grasp scenarios. Specifically, our design enhances grasping efficiency by enabling automatic switching between sensing modalities and optimizing the timing of grasp execution, while also achieving precise differentiation of visually similar objects.
 
\subsubsection{Robust Distance Measurement Across Different Targets}
\begin{figure*}[htbp]
    \centering
    \includegraphics[width=\textwidth,keepaspectratio]{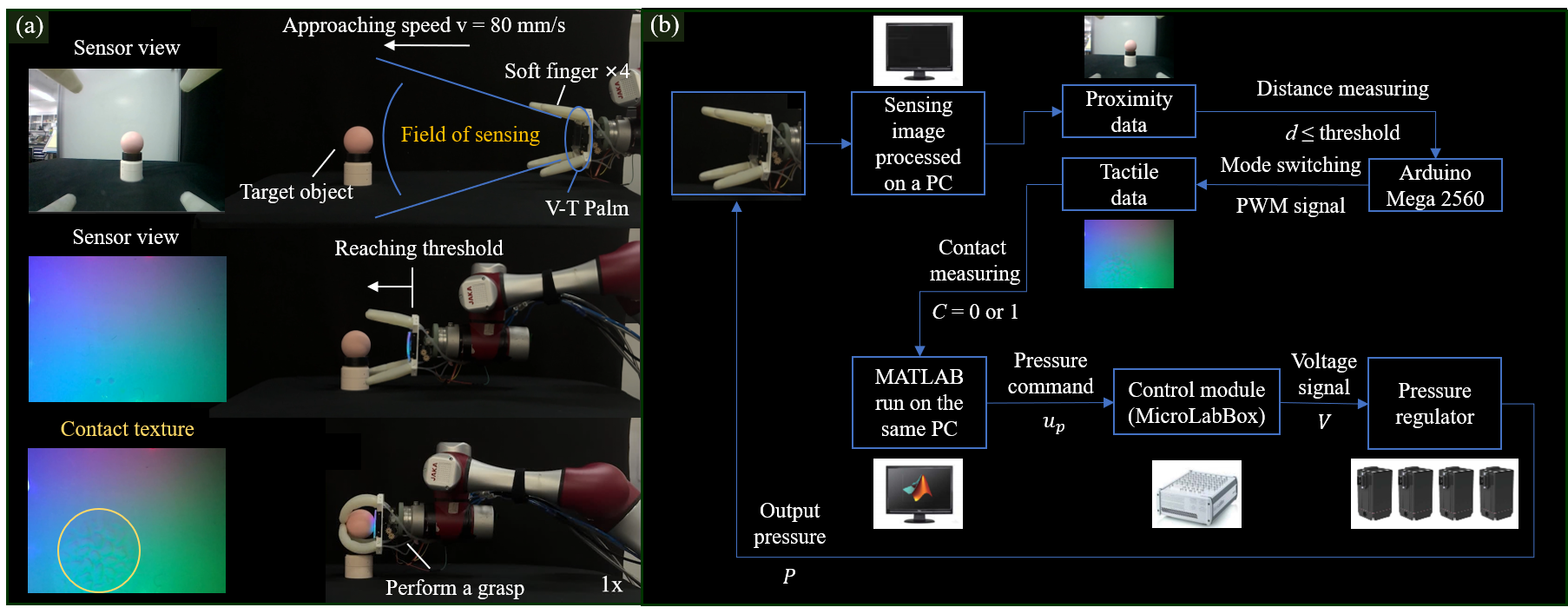}
    \caption{Schematic illustration of the operational Workflow and experimental platform. (a) shows the steps of the complete experimental process and the images from sensor perspective. (b) explains the control logic of the platform.}
    \label{fig:Appli_exp_1}
\end{figure*}

% \subsubsection{Proximity perception of further distance}
As shown in Fig.~\ref{fig:Appli_exp_1}(a), the V-T Palm was integrated with four neumatic soft fingers to form a sensor-actuator-integrated flexible gripper, which was mounted on a 6-DOF robotic arm (Jaka Zu7, JAKA). The target object was placed stationary on a horizontal tabletop, with the measuring surface of the palm oriented vertically. Under the proximity mode, the palm was mechanically coupled with the end of the robotic arm and approached the target at a constant speed of 80 mm/s. For different target objects, the segmentation mask used in the ranging algorithm required re-calibrated. When the measured distance reached 10 cm, the sensor modality switching procedure was triggered. Upon detecting definitive contact, the grasping procedure was then initiated.

\begin{figure*}[htbp]
    \centering
    \includegraphics[width=\textwidth,keepaspectratio]{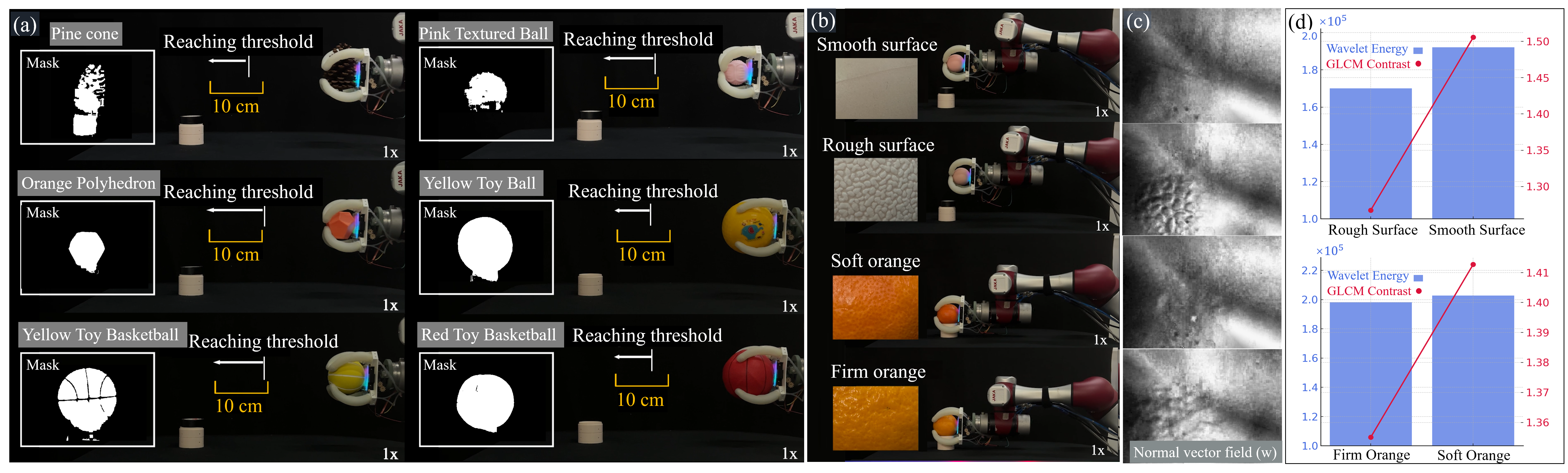}
    \caption{Experimental results of pre-planned grasping and subtle identification tests. (a) shows the grasping performance for different objects. (b), (c) and (d) explains the reconstruction results and roughness analysis of the tactile palm.}
    \label{fig:Appli_exp_2}
\end{figure*}

The input pressures of each soft actuator are regulated by an electronic pressure regulator (QB1XANEEN100P400KPG, Proportion-Air), which are connected to a control module (MicroLabBox-DS1202, dSPACE). The sensor transmits image data to the PC via a serial port, while a control module (Arduino Mega 2560, Arduino SRL) manages communication between the sensor and the PC. The system continuously monitors the sensor's image signals in real time to execute logical decisions. Upon the initiation of a grasp command, MATLAB generates and transmits control signals at a frequency of $F_{s} = 50$ Hz (shown in Fig.~\ref{fig:Appli_exp_1}(b)).

In this experiment, the target object's position and the robot arm's motion trajectory were fixed. We evaluated the success rate of the bimodal hand in autonomously sensing modalities switching and executing grasps for different targets, achieving a 100\% grasping success rate and a 81.8\% ranging accuracy, as presented in the Supplementary Video and Fig.~\ref{fig:Appli_exp_2}(a). 

\subsubsection{Distinguishing Visually Similar Objects}
In high-speed sorting applications, distinguishing target objects with highly similar visual characteristics (e.g., color and shape) poses a significant challenge, particularly when relying solely on vision sensors, which may struggle with subtle texture variations or lighting inconsistencies. To demonstrate the effectiveness of our approach, we conducted controlled experiments with two test groups: pink yoga balls featuring different surface textures and oranges at various stages of ripeness. As shown in Figs.~\ref{fig:Appli_exp_2}(b-d), for each reconstructed result, we applied the wavelet transform to each reconstructed result to compute the energy (sum of squares) of wavelet decomposition coefficients across different scales. Additionally, the gray-level co-occurrence matrix (GLCM) was utilized to quantify grayscale contrast variations. The experimental results demonstrate that our sensing system can effectively differentiate target objects, even when their visual attributes are highly similar, highlighting its potential for improving high-speed sorting performance.

\subsection{Card Insertion Experiment}

\begin{figure}[!t]
    \centering
    \includegraphics[width=\columnwidth,keepaspectratio]{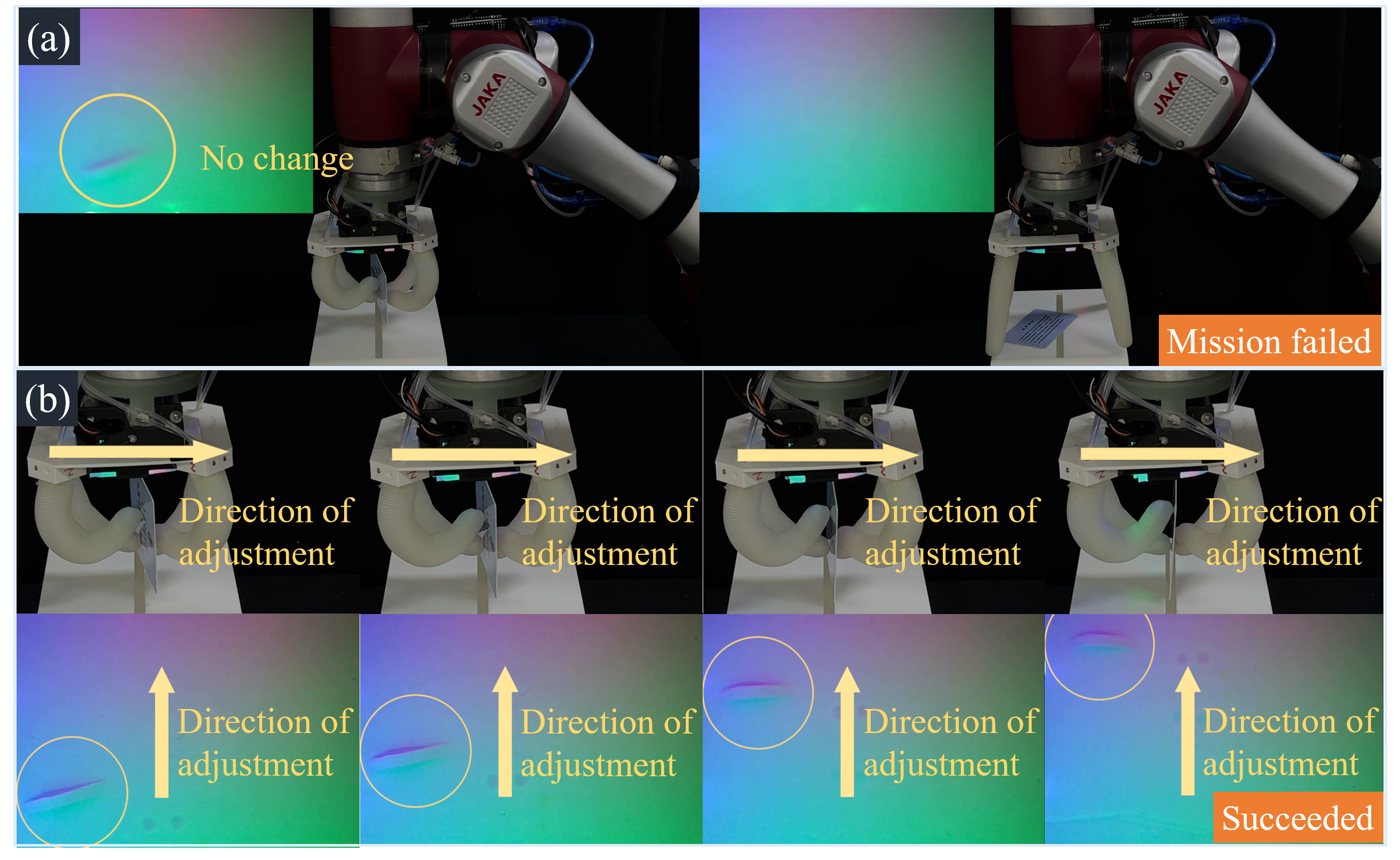}

    \caption{Display the use of V-T Palm's DOF in hand fine-tuning card delivery application. (a) shows the failure case of direct throw. (b) demonstrates a successful process based on feedback of tactile information.}
    \label{fig:Appli_exp_3}
\end{figure}

The conveyor-driven transmission mechanism naturally endows the haptic palm with one degree of freedom in movement. Leveraging this capability, we can fine-tune the target's orientation without altering the grasping state of the fingers. As shown in Fig.~\ref{fig:Appli_exp_3} and the supplementary video, in the `card transfer' experiment, when the robotic arm’s end effector is aligned with the upper edge of the `transfer gap', simply releasing the hand grip to perform a throwing action does not guarantee that the card will successfully pass through the narrow gap. By relying solely on tactile feedback from the palm and its ability to adjust along a single degree of freedom, the card's orientation can be dynamically refined without changing the robotic arm’s joint configurations or modifying the grasping control program. The successful execution of the throwing task offers a new possibility for soft robotic hand dexterous manipulation.

%%%%%%%%%%%%%%%%%%%%%%%%%%%%%%%%%%%%%%%%%%%%
%%%%%%%%%%%%%%%%%conclusion%%%%%%%%%%%%%%%%%%%%
%%%%%%%%%%%%%%%%%%%%%%%%%%%%%%%%%%%%%%%%%%%%
\section{CONCLUSION}

In this paper, we propose a visual-tactile dual-mode palm sensor with a sliding sensing window. Through the introduction of mechanical transmission design, the sensor can seamlessly switch between vision-based ranging perception and tactile-based texture perception. The developed ranging algorithm enables accurate distance sensing across different motion speeds. In tactile mode, the multi-layer neural network is employed to achieve the fine texture perception with ultra-high altitude resolution. Experimental results demonstrate that our proximity perception system can accurately detect distances within 0–50 cm, achieving its highest accuracy when the target is moving at 17.5 cm/s. The average tracking error is 0.183 in the image coordinate system. The tactile mode provides a perception density 6–8 times that of human fingertip skin. Also, realized high-precision three-dimensional shape reconstruction of the contact surface.

The proposed design offers significant advantages in adaptive robotic grasping, object sorting, and fine-tuned hand adjustments. By integrating the dual-mode palm sensor with soft fingers, the robot can rapidly and reliably grasp objects of varying shapes at arbitrary speeds, achieving a 100\% success rate. Furthermore, the sensor enables fast classification of visually indistinguishable objects by reconstructing their surface texture morphology, allowing for differentiation of roughness and material properties, such as the ripeness of fruit (soft vs. hard). Through the 'card throwing' experiment, the effectiveness of the intrinsic palm DOF in enabling in-hand fine-tuning operations has been validated.

At the same time, the proposed design method ensures complete non-occlusion of the sensor's lens during proximity measurements. This feature enables the sensor to facilitate pre-planning of the target grasping posture, which will be further explored in our future work. 

%%%%%%%%%%%%%%%%%%%%%%%%%%%%%%%%%%%%%%%%%%%%
%%%%%%%%%%%%%%%%%ACKNOWLEDGMENT%%%%%%%%%%%%%%%%%%%%
%%%%%%%%%%%%%%%%%%%%%%%%%%%%%%%%%%%%%%%%%%%%

\section{ACKNOWLEDGMENT}
The author thanks Ms. Jiawen Yu for her suggestions on the experimental design of this article and to Mr. Xinyu Yang for his support in data processing.

\begingroup
\footnotesize
% \bibliographystyle{IEEEtran}
% \bibliography{refs}
\input{main_version6_arxiv_0411.bbl}

\endgroup

\end{document}